\documentclass[english,preprint]{elsarticle}
\usepackage[T1]{fontenc}
\usepackage{url}
\usepackage{graphicx}
\usepackage[commentsnumbered,ruled,vlined]{algorithm2e}
\usepackage{comment}
\usepackage{enumitem}
\usepackage{xcolor}
\usepackage{verbatim}
\usepackage{amsmath,amssymb,amsfonts}
\usepackage{algorithmic}
\usepackage{graphicx}
\usepackage{textcomp}
\usepackage{caption}
\usepackage{subcaption}
\usepackage{booktabs}
\usepackage[utf8]{inputenc}
\makeatletter

\usepackage{amssymb,amsmath,amsthm}

\journal{ArXiv}

\begin{document}

\begin{frontmatter}

\title{Using Wavelet Domain Fingerprints to Improve Source Camera Identification}
\author[1]{Xinle Tian\corref{cor1}} \ead{xt373@bath.ac.uk}
\author[1]{Matthew A. Nunes} 
\cortext[cor1]{Corresponding author}
\author[1]{Emiko Dupont} 
\author[2]{Shaunagh Downing}
\author[2]{Freddie Lichtenstein}
\author[2]{Matt Burns}
 \affiliation[1]{                   
addressline={Department of Mathematical Sciences, University of Bath, Claverton Down},                  
postcode={BA2 7AY},                   
city={Bath},                   
country={United Kingdom}}
 \affiliation[2]{organization={CameraForensics}, 
city={Bristol},                  
postcode={BS1 6AA},                   
country={United Kingdom}}

\begin{abstract} 

Camera fingerprint detection plays a crucial role in source identification and image forensics, with wavelet denoising approaches proving particularly effective for extracting sensor pattern noise (SPN). In this article, we introduce the concept of a wavelet domain (WD) fingerprint, redefining the representation of the extracted fingerprint from the conventional image domain to the native wavelet coefficient domain. Rather than reconstructing the fingerprint as a spatial domain image, fingerprint comparison is performed directly on the wavelet coefficients, eliminating the final inverse transform and subsequent image-domain post-processing. This reformulation streamlines the fingerprint extraction and comparison pipeline while preserving the information required for source camera identification. The proposed framework is applicable to existing wavelet-based SPN extraction methods and is demonstrated using two representative state-of-the-art pipelines. Experimental results on real-world datasets show that the proposed approach significantly reduces computational cost, making it well-suited for large-scale source camera identification applications.

\end{abstract}


\end{frontmatter}

\begin{center}
\textbf{Published in \\
\textit{Forensic Science International: Digital Investigation}\\
https://doi.org/10.1016/j.fsidi.2026.302196
}
\end{center}

\section{Introduction}
Camera fingerprint detection is a fundamental technique in digital image forensics, aimed at determining the source device of an image and verifying its authenticity, while detecting any manipulations applied to them \cite{rocha2011vision}. The analysis of digital images or videos can serve as scientific evidence in legal cases involving surveillance, cybercrime, and other forensic investigations, for example, to combat film piracy or fake news generation. One key area of digital image forensics is source camera identification (SCI), which seeks to determine the specific camera used to capture an image. This plays a crucial role in forensic inquiries, aiding in the verification of image origins and supporting legal proceedings \cite{garfinkel2010digital, redi2011digital}.

Common methods for source camera identification rely on various image characteristics, including lens aberration \cite{choi2006automatic}, sensor noise patterns \cite{lukas2006digital}, chromatic aberration \cite{van2007identifying}, CFA interpolation \cite{bayram2005source, cao2009accurate, chen2015camera}, as well as classification models based on features extracted from images using classical statistical measures \cite{kharrazi2004blind} or more recently automatically via deep learning (see e.g. \cite{huang2018identification, wang2018source, freire2019deep, mandelli2020cnn, lu2024source}).  With advancements in digital forensics, researchers have established that each camera produces a unique fingerprint within its images, commonly known as sensor pattern noise (SPN), which is dominated by Photo-Response Non-Uniformity (PRNU) \cite{bartlow2009identifying, manisha2022beyond}. There has been a growing literature on PRNU-based device identification, see e.g. \cite{lukas2006digital, goljan2008camera, goljan2009large, akshatha2016digital, perez2016fast} and PRNU image clustering, see e.g. \cite{marra2017blind, khan2019fast, jiang2020blind}.  This method of source identification is considered highly reliable \cite{kang2011enhancing, gisolf2013improving}, as PRNU arises from manufacturing imperfections in camera sensors, making it distinct for each model \cite{albisani2021checking}.  For overviews of PRNU extraction methods, we refer the reader to \cite{kirchner2015forensic, al2016spn}.

Among PRNU-based camera source identification methods, frequency domain denoising approaches, using Fourier and wavelet transforms, have gained attention for their effectiveness in isolating high frequency noise components (believed to contain the camera sensor pattern noise) from structured image content, see e.g. \cite{lukas2006digital, chen2008determining, amerini2009analysis, zeng2020source}. In this process, a sensor pattern fingerprint is extracted from each image using a combination of image transforms and frequency domain filtering. These components are often subsequently processed or additionally filtered in the image domain to remove unwanted artifacts before comparison, see e.g. \cite{chen2008determining, gloe2012unexpected, al2016spn}. More recently, numerous deep learning-based SCI methods have been proposed in the literature, see for example \cite{mandelli2020cnn, karunakar2021identification, xiao2022effective, bharathiraja2023deep, lu2024source, han2025dwt}. 

\subsection{Motivation for our work}

In many scientific applications, such as cybercrime, forensic analysts and crime agencies are faced with the challenging scenario in which they have a potentially suspect candidate image and it is of interest to compare this image to all entries of a large image database to determine possible camera matches.  In this setting, it is rare to have sufficient copies from the same camera on which to train a full deep learning classification model, precluding the use of such methods for camera identification, and thus an unsupervised approach is needed; see also \cite{goljan2009large} for additional insight on the need for large-scale testing in digital forensics.  

Our research team includes industry practitioners, whose operational requirements and regulatory considerations motivates the use of transparent and interpretable methodologies. Outside the deep learning methods for PRNU comparison listed above, some authors have also proposed explainable AI-based (XAI) techniques (see e.g. \cite{nayerifard2026source}). Whilst Interpol recognises potential in these techniques \cite{geradts}, there is significant skepticism and mistrust of AI-based methods in digital forensics \cite{solanke} and they still remain difficult to explain in judicial proceedings \cite{Chaudhary}.  A recent Interpol review notes that since the deep learning PRNU methods are inherently less transparent, there is a critical need for rigorous validation and standardized reporting when using these techniques \cite{geradts}. Indeed, international courts have thusfar employed (legally admissible) PRNU-based evidence using traditional (non-AI) techniques \cite{mackenzie2018photo}. 

Consequently, the proposed framework in this article is deliberately developed without relying on machine learning techniques, instead focusing on classical statistical / image processing approaches that provide greater explainability and facilitate regulatory acceptance in practical forensic applications.  Motivated by the intended industrial application, we particularly focus on the situation in which a large number of image comparisons is to be undertaken; it is of the utmost importance that the image processing involved in such comparisons is computationally efficient.  As such, the primary contribution of our work is to develop a fast and efficient unsupervised SCI framework for the non-pretraining vast-comparison setting. We note here that the frequency domain denoising methods considered in this article may not be competitive with deep learning methods in classification tasks when multiple reference images are available, but we specifically aim to address scenarios in which no reference fingerprint of the source camera is available, and images must be compared directly to large image corpora. In this setting, practitioners are willing to sacrifice a small amount of accuracy for computational speed since the goal is to identify potential source relationships for further investigation.

In this article, we propose a general approach to modify existing state-of-the-art frequency-based fingerprint extraction and processing frameworks to improve the computational efficiency of PRNU-based image comparison while maintaining or improving detection accuracy. To achieve this, we advocate performing fingerprint comparison directly in the wavelet domain, rather than following the conventional approach of reconstructing the extracted PRNU component in the image domain prior to comparison. The proposed modification is motivated by the orthonormality of the Fourier and wavelet transforms \cite{nason1994discrete}, implying that the information contained in the filtered frequency-domain coefficients is equivalent to that in the reconstructed image-domain fingerprint for the purpose of similarity comparison. Consequently, the inverse wavelet transform and subsequent image-domain post-processing can be eliminated, streamlining the extraction pipeline and reducing computational cost while preserving the information required for fingerprint comparison. These computational savings are particularly beneficial for the large-scale comparison scenarios described above. It should be noted that the objective of this work is not to develop new denoising filters, but rather to investigate whether existing wavelet-based PRNU extraction frameworks can be made more computationally efficient through direct WD fingerprint comparison. 

We exemplify our proposed framework with two widely adopted wavelet-based PRNU extraction frameworks: the Locally Adaptive Window-based (LAW) denoising pipeline and the recently proposed Dual-Tree Complex Wavelet Transform (DTCWT) approach, and proposes wavelet-domain variants of each. The proposed framework is evaluated according to (i) classification performance, quantified using the Area Under the ROC Curve (AUC) and the True Positive Rate (TPR) at a fixed True Negative Rate (TNR), (ii) computational simplicity, measured by the average fingerprint extraction time, and (iii) in-use computational efficiency, as measured by the time taken for fingerprint comparisons; for completeness we also record the total end-to-end processing time for undertaking a set of comparisons. 

\subsection{Structure of the article}
The rest of the article is organised as follows.  Section \ref{sec:background} provides a brief overview of wavelet transforms and details common denoising filters used in PRNU extraction.  Our proposed approach to modify these denoising filters is introduced in Section \ref{sec:Method}, with experiments demonstrating the pipeline in Section \ref{sec:experiment}.  Section \ref{sec:Conclusion} concludes and provides a discussion of the contribution of this article.

\section{Background}
\label{sec:background}

\subsection{Discrete wavelet decompositions in image denoising}
\label{sec:denoising}

Prior research has posited that images captured by a camera can be conceptualised as a combination of the "true scene" of an image, coupled with a camera fingerprint and additional noise \cite{chen2008determining}. Typically, the representation of the actual scene is retained within the low frequency components of the image data, whereas the noise/fingerprint is contained in the higher frequency components.  This motivates the use of wavelet transforms, a natural tool for decomposing images into different frequency components.  

A wavelet transform represents an input image as a linear combination of information in different directions and at different frequencies (known as wavelet scales or levels) up to a given level $J$, known as the primary resolution level.  In practice, the coefficients of the transform are obtained from successive applications of low-pass and high-pass filters on the rows and columns of the image, resulting in direction-specific coefficients for the horizontal, vertical and diagonal features at each level.  Starting from the finest resolution of the image, the low-pass filter produces `approximation' or `scaling' coefficients corresponding to a coarser representation of the image, while the high-pass filter outputs `detail' coefficients representing the details lost in this approximation. The detail coefficients obtained through this process together with the approximation coefficients at the coarsest resolution $J$ form a pyramid structure in which the total number of coefficients equals the image size. The transformation from the image domain to these coefficients in the wavelet (frequency) domain is an invertible linear transformation, referred to as the discrete wavelet transform (DWT).

\begin{figure*}[!h]
\centering
\includegraphics[width = \linewidth]{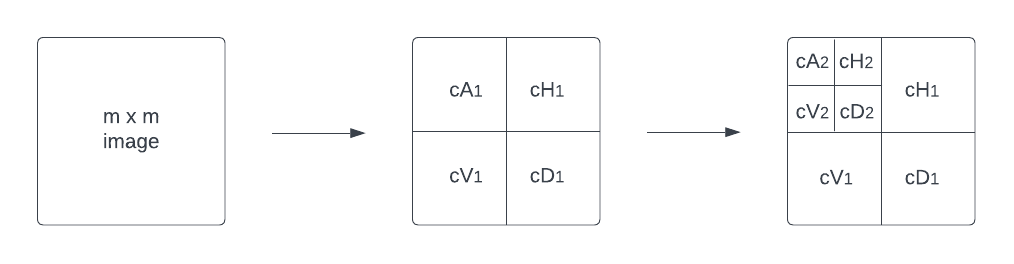}
\caption{Illustration of DWT applied to an image with level $J = 2$.}
\label{fig:DWT}
\end{figure*}

The output of the DWT is often visually presented as a subdivision of the original image domain into a collection of smaller subimages, each of which illustrate different components of the transform. 
More specifically, the approximation coefficients (labelled $cA$) correspond to a subimage capturing the lowest resolution features of the original image such as the background scene or mean signal, while for each level of the transform, the detail coefficients (denoted $cH$, $cV$, $cD$ for the horizontal, vertical and diagonal directions, respectively) lead to three subimages showing the details that would be added in each direction if the resolution of the image approximation was increased by one level.  An illustration of this process can be seen in Figure \ref{fig:DWT}.   

In image denoising, the DWT is combined with thresholding or shrinking of the finer scale detail coefficients in the wavelet domain. This will suppress the high frequency components of the image data most likely to be noise, while keeping the remaining features of the image intact. 
The denoised image is then reconstructed from the modified coefficients using the inverse DWT.  For comprehensive overviews of wavelet transforms and denoising, we refer the reader to \cite{nason2008wavelet, vidakovic2009statistical, sundararajan2016discrete}.  

The SPN is often seen as the residual between the noise-suppressed image and the random background noise; this can be obtained either through retaining certain high frequency components in the wavelet domain before inversion \cite{chen2008determining} or by computing the residual image in the original data domain \cite{zeng2020source}.  However, SPN extraction is challenging as the unstructured high frequency behaviour of the SPN makes it difficult to distinguish it from random noise in the image. Whilst classical wavelet thresholding and coefficient shrinkage methods are well-established in the statistical literature for signal denoising, they do not perform well in the SPN setting.  This has led researchers to explore spatially-adaptive denoising, such as estimating the noise field across the image to inform filtering methods, such as Wiener filtering, see e.g. \cite{mihcak1999spatially}.  In addition, other wavelet transforms have been used recently to improve spatial or directional information used in the denoising, at the expense of possible increased computational cost, such as undecimated wavelet transforms \cite{argenti2006mmse}, or the dual tree complex wavelet transform (DTCWT) \cite{zeng2020source}.

\subsection{Existing PRNU denoising methodology}
\label{sec:existing}

In this section, we provide details on the main two PRNU extraction methods used later in the article, which we modify to demonstrate and evaluate our proposed WD fingerprint methodology. 

More specifically, we describe 1) the \textit{Locally Adaptive Window-based denoising} (LAW) method that has been proposed by \cite{mihcak1999low} and further studied in the PRNU context by \cite{lukas2006digital, chen2008determining, junior2019prnu}, as well as 2) the denoising approach based on the dual tree complex wavelet transform of \cite{zeng2020source}.  We focus on these two wavelet filtering methods since the LAW method is one of the most commonly-used in the literature, and the method of \cite{zeng2020source} has been shown to outperform other state-of-the-art PRNU extraction methods such as the recently proposed BM3D filtering technique \cite{cortiana2011performance}, known also to be computationally intensive to run.

\subsubsection{Locally Adaptive Window-based denoising (Mihcak's filter)}
\label{sec:DWT}

The algorithm proposed in \cite{lukas2006digital} is a wavelet-based denoising method that uses local statistical modelling of wavelet coefficients.  The spatially-adaptive filters used in this SPN workflow is referred to as \textit{Mihcak’s filter} in some previous studies \cite{amerini2009analysis}. The proposed SPN extraction method begins by applying the DWT to the noisy image ($I_{m \times m}$), decomposing it into subimages at multiple scales and orientations. 
Since the coefficients in each subimage are naturally spatially arranged due to being (smaller) filtered versions of the original image, instead of using a global threshold for coefficient shrinkage, the method estimates the local mean and variance of high frequency wavelet coefficients within small neighbourhoods, assuming a local Gaussian distribution. Specifically, for each coefficient $ w(k_1,k_2) $, the local variance is estimated using a neighbourhood window $ \mathcal{N}(k_1,k_2) $ around the coefficient:
\begin{equation} \label{equ:var_est}
    \hat{\sigma}_N^2(k_1,k_2) = \max\left(0, \frac{1}{N} \sum_{(i_1,i_2) \in \mathcal{N}(k_1,k_2)} \hspace{-0.7cm}w^2(i_1,i_2) - \sigma_n^2 \right),
\end{equation}
where $N$ is the number of coefficients in the local window, and $\sigma_n^2$ is the noise variance  of the image. 
The subtraction of $ \sigma_n^2 $ ensures that only the true local variance of the signal is retained. The application of the maximum function prevents negative variance estimates, which could otherwise occur in regions dominated by noise.  In practice, $\hat{\sigma}_N^2(k_1,k_2)$ is computed for a range of local windows, for example with $N=3^2,5^2,7^2,9^2$, and for each pixel, the final estimate $\hat{\sigma}^2(k_1,k_2)$ is taken to be the minimum of those computed  (see e.g.\cite{mihcak1999low, mihcak1999spatially, lukas2006digital}). 

Once the local variance is estimated, an \textit{adaptive local Wiener-like shrinkage} is applied to attenuate noise while preserving significant image structures. The denoised coefficient is computed as:
\begin{equation}
    \hat{w}(k_1,k_2) = \frac{\hat{\sigma}^2(k_1,k_2)}{\hat{\sigma}^2(k_1,k_2) + \sigma_n^2} w(k_1,k_2).
\end{equation}

This equation scales each coefficient based on the ratio of its local signal variance to the total variance (signal + noise). When the local variance $ \hat{\sigma}^2(k_1,k_2) $ is much larger than the noise variance $ \sigma_n^2 $, the shrinkage factor approaches 1, meaning the coefficient remains largely unchanged. However, when the local variance is close to the noise variance, the shrinkage factor becomes small, reducing the coefficient’s magnitude to suppress noise. This adaptive process ensures that edges and textures in the image are preserved while noise is attenuated more aggressively in smooth areas, leading to effective denoising without excessive blurring.

The filtered coefficients after the shrinkage are then transformed back to the image domain using the inverse wavelet transform (IDWT) to reconstruct the filtered image ($\tilde{I}_{m \times m}$). For colour (multi-channel) images, this procedure is done to each image channel, and the resulting channels are often combined using a luminance-based grayscaling using the linear combination of the R, G, B channels (see e.g. \cite{goljan2008camera, al2016spn}):
$$\tilde{I}_{m \times m}=0.299\tilde{I}^R_{m \times m}+0.587\tilde{I}^G_{m \times m}+0.114\tilde{I}^B_{m \times m}.$$

\begin{figure*}[!h]
\centering
\includegraphics[width = \linewidth]{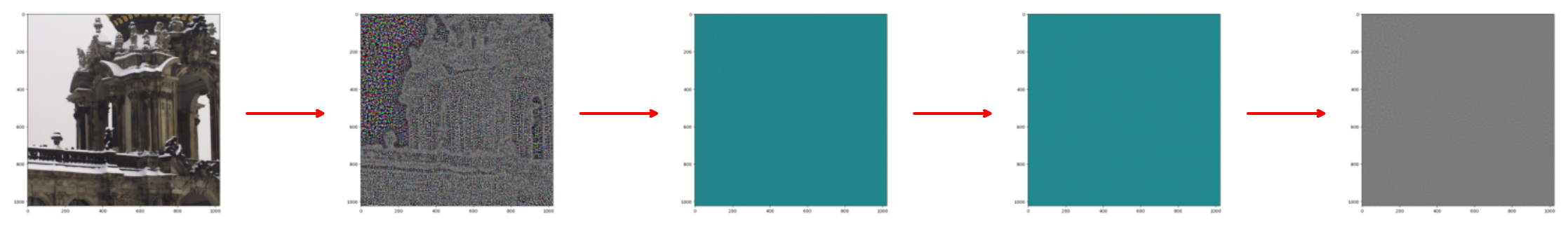}
\caption{Illustration of the LAW method applied to an image. The PRNU extraction image processing pipeline consists of (i) wavelet decomposition with spatially adaptive filtering and inversion (ii) grayscaling of denoised colour images (iii) zero-mean normalisation and (iv) second Wiener filtering to extract the fingerprint.}
\label{fig:LAW}
\end{figure*}

\textit{Denoised image post-processing.} In PRNU applications, the resulting denoised images $\tilde{I}_{m \times m}$ may still contain geometric artifacts caused by the sensor design, which it is recommended to remove prior to comparison \cite{chen2008determining}; these artifacts are sometimes referred to as \textit{shared components} \cite{al2016spn}.  Typically, this artifact removal consists of two steps.  Firstly, row- and columnwise means are subtracted from the denoised image; this \textit{zero-mean} operation helps in removing linear patterns found in lower quality cameras \cite{chen2008determining}.  A final \textit{Wiener-type filtering} step is performed in the Fourier domain, to further refine the PRNU estimate by suppressing any residual polygonal structures in the image. This is achieved by computing the 2D Discrete Fourier Transform (DFT) to transform the image $\tilde{I}_{m\times m}$ into the frequency domain:
\begin{equation}\label{eq:DFT}
    F(\omega_1,\omega_2) = \mathcal{F}\{ \tilde{I}_{m\times m} \}(\omega_1,\omega_2),
\end{equation}
where $F(\omega_1,\omega_2)$ represents the Fourier frequency coefficients of the noisy image. Adaptive Wiener filtering is then applied to the Fourier coefficients, where the Wiener filter estimates the power spectrum of the noise and applies a signal-dependent attenuation:
\begin{equation}\label{eq:DFTsa}
    \hat{F}(\omega_1,\omega_2) = F(\omega_1,\omega_2) \cdot \frac{\hat{S}(\omega_1,\omega_2)}{\hat{S}(\omega_1,\omega_2) + S_n},
\end{equation}
where \( \hat{S}(\omega_1,\omega_2) \) is the estimated local power spectrum of the image $\tilde{I}_{m\times m}$, computed using the same method as described in Equation \eqref{equ:var_est} and \( S_n \) is the estimated power spectrum of the noise. Lastly the filtered Fourier coefficients are converted back into the image domain using the inverse DFT (IDFT):
\begin{equation}\label{eq:iDFT}
f(k_1,k_2) = \mathcal{F}^{-1}\{ \hat{F}\}(\omega_1,\omega_2).
\end{equation}
This results in a denoised fingerprint $f_{m\times m}$ where remaining noise distortions have been adaptively suppressed. A detailed algorithm for LAW is presented in the pseudocode in Algorithm \ref{alg:LAW} for the general case of coloured images. Figure \ref{fig:LAW} illustrates the LAW method process for extracting camera fingerprints using an image from an open data source \cite{gloe2010dresden}, which can be accessed from \textit{https://www.kaggle.com/datasets/micscodes/dresden-image-database}.

\begin{algorithm}[!h]
\SetKwData{Left}{left}
\SetKwData{This}{this}
\SetKwData{Up}{up}
\SetKwFunction{Union}{Union}
\SetKwFunction{FindCompress}{FindCompress}
\SetKwInOut{Input}{Input}\SetKwInOut{Output}{Output}
\SetKwComment{comment}{\#}{}
\Input{Image $I_{m \times m}$}
\Output{SPN fingerprint $f_{m \times m}$}
\BlankLine

\textbf{STEP 1:}  Load image $I_{m \times m}$ \\

\textbf{STEP 2:}  Perform wavelet domain filtering:\\
\For{each image channel $i \in\{R, G, B\}$}{
\begin{enumerate}
\item Perform DWT with primary resolution level $J$\\
\item \SetAlgoNoLine \For{each wavelet level $j=1,\dots, J$}{
\begin{quote}Apply locally adaptive Mihcak filter to high frequency coefficients $cH, cV,cD$\\
\end{quote}     }
\item Set the approximation coefficients $cA$ to zero\\
\item Perform IDWT using zeroed $cA$ and filtered\\ coefficients $cH, cV,cD$ to get filtered image\\ channel $\tilde{I}^{i}_{m \times m}$\\
\end{enumerate}
}
    \textbf{STEP 3:} Perform grayscaling to combine filtered image channels\\
\textbf{STEP 4:} Apply Wiener filtering in Fourier domain: \\
\begin{enumerate}
        \item Perform DFT to obtain frequency components $F(\omega_1,\omega_2)$\\
        \item Apply locally adaptive Mihcak filter to obtain\\ filtered frequency components $\hat{F}(\omega_1,\omega_2)$ \\
        \item Perform IDFT to get final image domain fingerprint $f_{m \times m}$\\
        \end{enumerate}
 \caption{The LAW SPN fingerprint extraction algorithm for colour images}
 \label{alg:LAW}
\end{algorithm}

\subsubsection{Dual tree complex wavelet transform PRNU extraction}
\label{sec:DTCWT}

The (2D) dual tree complex wavelet transform was introduced by \cite{kingsbury1998dual} as a tool to extract improved spatial and directional information from images compared to the standard DWT.  The DTCWT consists of two sets of filters which operate on the components of an image to form complex-valued wavelet coefficients. Due to the two filterbanks, at each level of the transform, the three directional coefficient subimages ($cH, cD, cV$) of the DWT are replaced by six subimages ($cH^1, cV^1, cD^1, cH^2, cV^2, cD^2$) -- three for each branch of the tree \cite{selesnick2005dual}.  \cite{zeng2020source} introduced an SPN extraction method based on the DTCWT which was shown to perform well in SPN applications, especially when edges are present in images.

The SPN extraction method proposed in \cite{zeng2020source} has a similar form to a traditional decomposition-threshold-inversion denoising method, except that due to the SPN application, interest lies in retaining the noise from the denoiser as a residual image, rather than the denoised image itself.  More specifically, the method firstly performs the DTCWT with a chosen primary resolution level, subsequently thresholding the six wavelet coefficient subimages by applying the local spatially adaptive filtering of \cite{mihcak1999low, mihcak1999spatially} on the \textit{(complex) magnitude} of the coefficients.  The inverse DTWCT (IDTCWT) is then performed on the resulting coefficient pyramid to produce a denoised image; the SPN is computed as the residual of the denoised image from the original image.  An overview of the DTCWT SPN extraction method is in Algorithm \ref{alg:DTCWT}.

\begin{algorithm}[!h]
\SetKwData{Left}{left}
\SetKwData{This}{this}
\SetKwData{Up}{up}
\SetKwFunction{Union}{Union}
\SetKwFunction{FindCompress}{FindCompress}
\SetKwInOut{Input}{Input}\SetKwInOut{Output}{Output}
\SetKwComment{comment}{\#}{}
\Input{Image $I_{m \times m}$}
\Output{SPN fingerprint $f_{m \times m}$}
\BlankLine

\textbf{STEP 1:}  Load image $I_{m\times m}$ \\
\textbf{STEP 2:}  Perform wavelet domain filtering:\\
\For{each image channel $i \in\{R, G, B\}$}{
\begin{enumerate}
\item Perform DTCWT with primary resolution level $J$\\
\item \SetAlgoNoLine \For{each wavelet level $j=1,\dots, J$}{
\begin{quote} Apply locally adaptive Mihcak filter to 
high frequency coefficients $cH^{1}, cV^{1}, cD^{1}$, $cH^{2}, cV^{2}, cD^{2}$
    \end{quote} 
    }
\item Perform IDTCWT using filtered high frequency coefficients $cH^{1}, cV^{1}, cD^{1}$, $cH^{2}, cV^{2}, cD^{2}$ and approximation coefficients $cA$ to get filtered\\ image channel $\hat{I}^{i}_{m \times m}$\\
\item Form channel-specific SPN image by computing residual from original image channel:
$\tilde{I}^{i}_{m \times m} = I^{i}_{m\times m}-\hat{I}^{i}_{m \times m}$
\end{enumerate}
}
    \textbf{STEP 3:} Perform grayscaling to combine filtered colour image channels and get final image domain fingerprint $f_{m \times m}$
 \caption{The DTCWT-based SPN fingerprint extraction algorithm  for colour images}
 \label{alg:DTCWT}
\end{algorithm}

The approach of \cite{zeng2020source} does not discuss including shared component removal as employed by \cite{lukas2006digital, chen2008determining} for the DWT-based SPN extraction method (see Section \ref{sec:DWT}); for the tampering localization application the authors considered, this may not have been necessary or beneficial.  However, we discuss possible inclusion of this aspect of fingerprint extraction in Section \ref{sec:experiment}. 

\subsection{Criteria for source identification}
\label{sec:Criteria}
Establishing criteria to determine whether images originate from the same device is crucial for making accurate final decisions. This step is significant, as it can directly impact the outcomes of the classification process. However, it is also a flexible component that ultimately depends on the analyst's decisions and specific application context.

In our study, we applied the simple cosine similarity test for its efficiency in measuring the angular distance between feature vectors, making it a widely used metric for data comparisons. For a pair of feature vectors $f_1$ and $f_2$ (representing extracted fingerprints from two images) their cosine similarity is given by:  
\begin{equation}\label{eq:cosine}
    \frac{f_1 \cdot f_2}{\|f_1\| \|f_2\|}
\end{equation}
where  $f_1 \cdot f_2$ is the dot product of the vectors, and $\|f_1\|$ and $\|f_2\|$ are their respective Euclidean norms. This measure provides a normalized similarity score between -1 and 1, where a higher value indicates greater similarity between the vectors.

After we obtain the cosine similarity score for each image pair, we need a threshold, $\lambda$, to determine whether the two images originate from the same source. If the similarity score falls below this threshold, the images are classified as coming from different sources; otherwise, they are considered to be from the same source.

A simple and effective method to determine this threshold can be derived from Receiver Operating Characteristic (ROC) curve analysis, a graphical representation used to evaluate the performance of a binary classification model. The ROC curve illustrates the trade-off between the true positive rate (TPR) and the false positive rate (FPR) at various benchmark value settings of a parameter.  The TPR measures the proportion of actual positives correctly identified by the method, while the FPR quantifies the proportion of actual negatives that are incorrectly classified as positives. In our case, we consider the ROC curve obtained by varying the value of $\lambda$.

\subsubsection*{Optimal threshold selection via Youden's Index}

For a given ROC curve, Youden's Index \cite{youden1950index} is the maximum difference between the TPR and the FPR. This difference should ideally be large, so a natural way to identify the optimal value for $\lambda$ is given by

%
\begin{equation}\nonumber
    \lambda_{YI} =\operatorname{argmax}_\lambda (\text{TPR}(\lambda) - \text{FPR}(\lambda)).
\end{equation}

This method helps in selecting a threshold that provides the best trade-off between correctly identifying same-source images while minimizing misclassification errors.
The downside is that it may give multiple potential values, making it challenging to select the most appropriate one. 
Careful evaluation and validation is needed in order to choose the value that best balances sensitivity and specificity. In general, from an application perspective, it is advisable to choose the value that yields the highest TPR. \\

\subsubsection*{Optimal threshold selection via high TNR}

An alternative approach for setting the classification criterion is to control the true negative rate (TNR) (the proportion of actual positives correctly identified by the method) which should ideally be large. More specifically, if our desired value for TNR is $R$, then since $\text{TNR}(\lambda)=1 - \text{TPR}(\lambda)$, the chosen threshold is the value of $\lambda$ for which $\text{TPR}(\lambda)=1-R$.

This method minimises the risk of mistakenly classifying images as being from the same source when they are not, while still achieving good chances of correctly identifying images that do originate from the same source. This approach aligns with scientists' objectives and practical operational needs, where minimising false positives is critical for maintaining the integrity and reliability of forensic decisions.

In the applications in Section \ref{sec:experiment}, we will present results that include both the criterion values and their corresponding TPR and TNR.\\

\section{Proposed methodology}
\label{sec:Method}

In this section, we present our proposed approach to fingerprint identification, which streamlines wavelet-based extraction pipelines by removing unnecessary processing steps. In many cases, accuracy and/or computational efficiency of comparisons can be gained by performing comparisons directly in the wavelet domain. 

Since the wavelet transformation is a linear operator based on an orthonormal wavelet basis, Parseval's theorem states that the signal energy is preserved when performing the transform (or equivalently its inverse), see e.g. \cite{chui1992introduction}.  Moreover, the cosine similarity measure used for comparisons is invariant under orthonormal transformations. Motivated by this, we hypothesise that, not only do the essential properties of the SPN image remain the same before and after performing the inverse wavelet transform, but the key information useful to SPN comparison is more effectively harnessed from the wavelet coefficients. In effect, this defines a new notion of camera fingerprint, namely a representation of the SPN in the wavelet domain, that we refer to as the WD fingerprint. 

Building on this idea, we propose modifications to both of the wavelet-based PRNU extraction methods outlined in Section \ref{sec:existing}.

\subsection{Proposed single channel WD fingerprint extraction with the DWT}
\label{sec:oursingledwt}

In this section, we propose a modified version of the original LAW method outlined in Section \ref{sec:DWT}, which we refer to as \textit{Wavelet Domain Locally Adaptive Window-based denoising} (WDLAW). Instead of constructing an SPN fingerprint in the image domain, WDLAW defines a fingerprint directly in the wavelet domain where the SPN information corresponds to the vector of detail coefficients ($cH, cV, cD$) from the DWT decomposition. This reduces the dimension of the fingerprint, decreasing computational overhead at the comparison stage of analysis.

The length of the WD fingerprint, denoted by $l$, in this case can be computed as
\begin{equation}\label{eq:ldef}
l=\sum_{j=1}^{J} \frac{3}{4} \times \left( \frac{m}{2^{j-1}} \times \frac{m}{2^{j-1}} \right) = \frac{3}{4} m^2\sum_{j=1}^{J} (2^{-J+1})^2,
\end{equation}
where $J$ is the primary resolution level of the DWT. For any choice of $J$, $l<m^2$, and for small $J$, $l$ approximates to $\frac{3}{4}m^2$. When $J = 1$, only the three level 1 detail subbands are retained, resulting in a fingerprint containing 75\% of the original image coefficients; as the decomposition level increases, additional detail subbands from coarser levels are incorporated into the fingerprint, causing its size to increase and gradually approach that of the original image. In other words, for $J\neq0$ the number of pixels used for the WD fingerprints is always less than the total number of pixels of the image, and for small $J$, only about 75\% of the pixels,
compared to the full $m^2$ for the LAW method. Note that since we are only using the wavelet coefficients in the fingerprint, we do not need to perform the zeroing of the approximation coefficients $cA$ in STEP 2-3) of Algorithm \ref{alg:LAW}. 


Another step reconsidered for efficiency purposes in the proposed techniques is the row-/columnwise zero-mean operation to suppress regular grid artefacts before applying the final Wiener-type filter. Since our objective is to perform fingerprint comparison directly in the wavelet domain, we omit this operation in the proposed framework. Within the scope of the datasets and evaluation metrics considered in this particular study, our experimental results indicate that omitting this step does not affect the final source camera identification performance. We note, however, that the zero-mean operation may remain beneficial in applications where the visual fidelity or quality of the reconstructed fingerprint is of primary interest.

\begin{figure*}[!h]
\centering
\includegraphics[width = 0.6\linewidth]{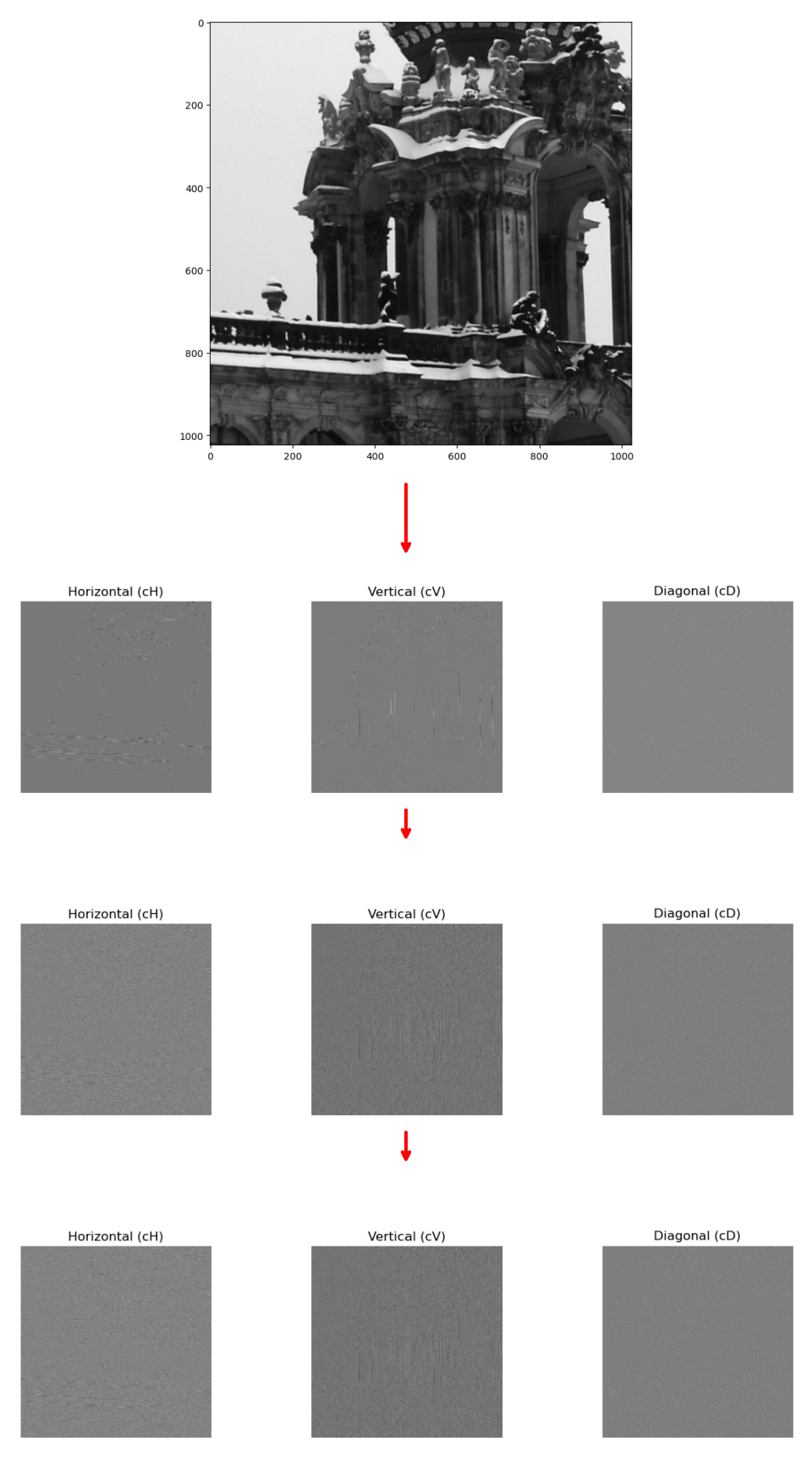}
\caption{Illustration of the WDLAW method applied to the grayscale image from Figure \ref{fig:LAW} with level $J = 1$, including (i) applying the DWT to obtain the high
frequency coefficients $cH, cV$ and $cD$ (ii) applying spatially adaptive filtering to these coefficients in the wavelet domain (iii) applying second Wiener filtering in the wavelet domain. The WD fingerprint is obtained by concatenating the resulting coefficients into a vector $f_l$. The columns from left to right correspond to $cH$, $cV$ and $cD$.}
\label{fig:gray}
\end{figure*}

An algorithmic description of WDLAW is shown in Algorithm \ref{alg:single-WDLAW}.  Note that due to the removal of the zero mean filtering and wavelet inversion steps, our proposed technique essentially performs Fourier domain filtering as described in \eqref{eq:DFT}-\eqref{eq:iDFT} immediately in the wavelet domain.  
Figure \ref{fig:gray} illustrates the WDLAW method for a single level wavelet decomposition for clarity. Note that in all figures, wavelet coefficients have been normalised for improved visualisation.  \\

\subsubsection*{Motivation for back-to-back filtering in the wavelet domain}

As described in the previous section, our primary motivation for removing the inverse DWT step in our modified PRNU extraction pipeline is computational and memory efficiency.  However, as noted above, this results in a "back-to-back" filtering procedure in which the Fourier-based Wiener processing is performed directly on the wavelet coefficients. 

Many authors have observed that SPN images are non-Gaussian, and that distributional changes are additionally induced by processing \cite{lukas2006digital, bondi2017design, fan2026blind}.  The assumption of independent and identically distributed Gaussian observations for Mihcak's filter will hence rarely hold in practice, especially since artifact removal steps are usually later in PRNU pipelines.  On the other hand, it is well-known that the wavelet transform has inherent decorrelation and Gaussianisation properties (see e.g. \cite{johnstone1997wavelet, nason2008wavelet}), resulting in closer adherence to this assumption of the filtering steps when operating directly in the wavelet domain.  On the datasets considered below, this filtering results in an improved classification performance, but this may not always be the case.

\begin{algorithm}[!h]
\SetKwData{Left}{left}
\SetKwData{This}{this}
\SetKwData{Up}{up}
\SetKwFunction{Union}{Union}
\SetKwFunction{FindCompress}{FindCompress}
\SetKwInOut{Input}{Input}\SetKwInOut{Output}{Output}
\SetKwComment{comment}{\#}{}
\Input{Image $I_{m \times m}$}
\Output{WD fingerprint $f_{l}$} \BlankLine

\textbf{STEP 1:}  Load image $I_{m \times m}$ \\

\textbf{STEP 2:}  Perform frequency domain filtering: 
\begin{enumerate}
\item Perform DWT with primary resolution level $J$\\
\item \For{each wavelet level $j=1,\dots, J$}{
\begin{enumerate}[labelindent=0mm]
\item Apply locally adaptive Mihcak's filter to high frequency coefficients $cH, cV,cD$\\
\item Apply Wiener filtering in Fourier domain: \\
\For{coefficient subimages $cH, cV,cD$}
{
\begin{enumerate}[labelindent=-2mm]
    \item Perform DFT to obtain frequency \\components $F(\omega_1,\omega_2)$\\
    \item Apply locally adaptive Mihcak filter to \\obtain filtered frequency components\\ $\hat{F}(\omega_1,\omega_2)$ \\
    \item Perform IDFT to get artifact-free\\ wavelet coefficients \\
\end{enumerate}
}
\end{enumerate}
}
\end{enumerate}
\textbf{STEP 3:} Concatenate wavelet coefficients into fingerprint $f_l$ 

\caption{The single channel WDLAW fingerprint extraction algorithm}
 \label{alg:single-WDLAW}
\end{algorithm}

\subsection{Proposed RGB image WD fingerprint extraction  with the DWT}
\label{sec:ourrgbdwt}

We now explore WD fingerprint extraction for RGB images.

As mentioned in Section \ref{sec:existing}, for colour images
grayscaling is usually performed after wavelet filtering and inversion in the PRNU processing pipeline (STEP 3 in both Algorithm \ref{alg:LAW} and Algorithm \ref{alg:DTCWT}).  As noted in \cite{al2016spn}, this grayscaling operation can instead be performed prior to analysis, so that the SPN extraction can be performed on a single `luminance' channel.  The motivation for the first approach stems from per-channel wavelet decompositions potentially providing more nuanced information on the channelwise SPN fingerprint (at the cost of three DWT computations), however this extra information is potentially subsequently `lost' when performing the grayscaling prior to artifact removal.  On the other hand, performing grayscaling first, whilst computationally more efficient, does not retain possibly useful colour information from the outset of the extraction.  The grayscaling operation will contribute to the computational cost of fingerprint extraction, but potentially speed up comparison of fingerprints.   An alternative is to keep all channels in the fingerprint, i.e. perform no grayscaling, at the cost of a more computationally expensive fingerprint comparison.

Our proposal for WD fingerprint extraction of colour images is to adopt a `grayscale first' approach, which we term gray-WDLAW.  In other words, this procedure begins by converting the input image to grayscale and then proceeds as in the single channel technique described in Algorithm \ref{alg:single-WDLAW}.   However, in the experimental evaluation in Section \ref{sec:experiment}, we also consider two other variants of our WDLAW technique for colour images (a) rgb-WDLAW, in which no grayscaling is performed, i.e. the fingerprint consists of wavelet coefficients from each direction and channel (of size $3l$, where $l$ is given by \eqref{eq:ldef}), and (b) WDLAW-gray, in which grayscaling is performed in the wavelet domain after the back-to-back wavelet and Fourier domain filtering.

\subsection{Proposed WD fingerprint extraction with the DTCWT}
\label{sec:ourdtcwt}

Given that SPN fingerprint extraction with the DTCWT in \cite{zeng2020source} is a relatively recent development, our interest also lies in testing whether a wavelet domain approach could be extended to use alternative wavelet decompositions beyond the traditional DWT. To this end, we consider several modifications to the extraction method of \cite{zeng2020source}.  Firstly, motivated by the LAW method, we introduce the LAW-DTCWT algorithm as our base extraction method, which employs the DTCWT at the decomposition and inversion stages in STEP 2 of Algorithm \ref{alg:LAW} instead of the DWT.  This differs from the `residual image' extraction approach of \cite{zeng2020source}, since it essentially performs low-pass filtering to remove mean structure in STEP 2-3) via setting approximation coefficients to zero; this could be viewed as a hybrid between the LAW method of \cite{lukas2006digital, chen2008determining} and that of \cite{zeng2020source}. Secondly, we propose a wavelet domain version of the LAW-DTCWT base technique.  More specifically, we perform the same modifications that we considered for the WDLAW method (grayscaling first, no wavelet inversion, no zero-mean operation), to form a WD fingerprint extraction method for RGB images, which we call gray-WDLAW-DTCWT. The detailed steps for applying our proposed wavelet domain approach are outlined in Algorithm \ref{alg:gray-WDLAW-DTCWT} below.

\begin{algorithm}[ht]
\SetKwData{Left}{left}
\SetKwData{This}{this}
\SetKwData{Up}{up}
\SetKwFunction{Union}{Union}
\SetKwFunction{FindCompress}{FindCompress}
\SetKwInOut{Input}{Input}\SetKwInOut{Output}{Output}
\SetKwComment{comment}{\#}{}
\Input{RGB Image $I_{m \times m}$}
\Output{WD fingerprint $f_{4l}$}
\BlankLine
\textbf{STEP 1:}  Load RGB Image $I_{m \times m}$ \\
\textbf{STEP 2:}  Perform grayscaling on $I_{m\times m}$\\
\textbf{STEP 3:}  Perform frequency domain filtering: 
\begin{enumerate}
\item Perform DTCWT with primary resolution level $J$\\
\item \For{each wavelet level $j=1,\dots, J$}{
\begin{enumerate}[labelindent=0mm]
\item Apply locally adaptive Mihcak's filter to high frequency coefficients $cH^{1}, cV^{1}, cD^{1}$, $cH^{2}, cV^{2}, cD^{2}$\\
\item Apply Wiener filtering in Fourier domain: \\
\For{coefficient subimages $cH^{1}, cV^{1}, cD^{1}$, $cH^{2}, cV^{2}, cD^{2}$}
{
\begin{enumerate}[labelindent=-2mm]
    \item Perform DFT to obtain frequency \\components $F(\omega_1,\omega_2)$\\
    \item Apply locally adaptive Mihcak filter to \\obtain filtered frequency components\\ $\hat{F}(\omega_1,\omega_2)$ \\
    \item Perform IDFT to get artifact-free\\ wavelet coefficients \\
\end{enumerate}
}
\end{enumerate}
}
\end{enumerate}
\textbf{STEP 4:} Concatenate wavelet coefficients into fingerprint $f_{4l}$ (of length $4l$, where $l$ is the length of the gray-WDLAW fingerprint)
 \caption{The gray-WDLAW-DTCWT fingerprint extraction algorithm for RGB images}
\label{alg:gray-WDLAW-DTCWT}
\end{algorithm}

Recall that for the DTCWT, the wavelet coefficients are complex-valued and hence are filtered according to their complex magnitude.  After the initial Mihcak filter, the resulting coefficients will still be complex-valued, and thus our wavelet domain approach means that the second Wiener filter needs to be modified for use on complex-valued coefficients.  We do this in a similar way to the first filtering, i.e. using the complex magnitude of the filtered coefficients.  Note that the resulting fingerprint from Algorithm \ref{alg:gray-WDLAW-DTCWT} will consist of six complex-valued subimages of coefficients before concatenation.  To perform fingerprint comparison with measures such as the cosine similarity \eqref{eq:cosine}, these coefficients need to be embedded into the real Euclidean space prior to computation of the metric.  In effect, due to \eqref{eq:ldef} this means that the fingerprints are vectors of size $4l$ ($2l$ coefficients, each of which has a real and an imaginary part).  This means that this approach requires more in terms of fingerprint storage and computational cost for fingerprint comparison compared with the WDLAW method, but may be beneficial in accuracy due to improved directional information \cite{kingsbury1998dual, selesnick2005dual, zeng2020source}. 

It is worth noting here that the DTCWT approach was originally developed for tampering localisation rather than fingerprint comparison. Therefore, its suitability for standalone source camera identification requires further investigation; the improved performance of the WD fingerprint modification of this method compared to that of the original is attributed to the double filtering performed akin to the LAW SPN extraction. 

Note that if a visual representation of the fingerprints is required, the WD fingerprints described in this section can be transformed to the image domain using the inverse wavelet transform.

\section{Evaluation of proposed PRNU extraction techniques}
\label{sec:experiment}

In this section, we describe the experiments undertaken to evaluate our proposed WD fingerprint extraction methods.  

\subsection{Experimental setup}
\label{sec:setup}

Our experiments were performed using RGB images from the open-source \textit{Dresden Image Database} \cite{gloe2010dresden}. For our analysis, we selected a subset of the dataset consisting of original JPG images from 26 different camera devices, with each device contributing 5 images. Each image was centre-cropped to a resolution of $1024 \times 1024$ pixels. Our objective is to perform pairwise comparisons of all images within this subset to determine whether two images originate from the same camera, a total of 8385 comparisons. 

Such large-scale comparisons are representative of practical industrial applications. By maintaining a controlled and uniform image size, the approach focuses on accurately identifying images from the same source among a large collection of images.

The DWT-based extraction method of \cite{lukas2006digital, chen2008determining} was performed using code from \textit{https://github.com/polimi-ispl/prnu-python}, a Python implementation of the Binghampton Matlab PRNU library \cite{binghamptonmatlab}; our proposed wavelet-domain algorithms were implemented using modifications to this code. The DTCWT-based PRNU extraction (in the original form of \cite{zeng2020source} as well as our proposed modifications to the technique), was implemented in Python based on Matlab code made available by the authors \cite{zengcode}.  We opted to use our own Python implementation of \cite{zeng2020source} to ensure as fair a comparison as possible of runtimes between the methods. 

In our comparison, we evaluated results using several key metrics: TPR and TNR based on the Youden index threshold (threshold 1 as described in Section \ref{sec:Criteria}), TPR when TNR is fixed at $R=0.99$ to minimize false identifications in real-world applications (threshold 2), area under the curve (AUC), and computational time in seconds, including extraction time per pair of images, comparison time per pair of images, and the total time for all comparisons. We compared four approaches to DWT-based extraction: the original LAW method, the grayscale first WDLAW method (gray-WDLAW), the wavelet domain grayscaling approach (WDLAW-gray), and the full RGB-channel WD fingerprint (rgb-WDLAW). Additionally, we considered four methods based on the DTCWT: the original method of \cite{zeng2020source}, as well as this DTCWT-based algorithm including additional image domain artifact removal (zero mean filtering, Fourier domain Wiener filtering) as described in Section \ref{sec:DWT}; we compared these methods with the hybrid-LAW version of DTCWT extraction (LAW-DTCWT) and our proposed gray-WDLAW-DTCWT method introduced in Section \ref{sec:ourdtcwt}. Our goal was to identify methods that achieve both high accuracy and efficient computation. 

To ensure a fair comparison, as far as possible all parameters involved in the estimation and computation of LAW-related methods were kept consistent, including the choice of wavelet, decomposition level, and noise variance. The recommended settings from the LAW method were applied, specifically using the Daubechies-4 (db4) wavelet and a primary decomposition level of $J=4$. We set the noise variance to be $\sigma_n =1.8^2$, following previous studies \cite{zeng2020source}. 

\subsection{Experimental results}
\label{sec:results}

\subsubsection{DWT-based extraction}

The results shown in Tables \ref{app:result1} and \ref{app:result2} demonstrate that, overall, our WDLAW methods outperform the original LAW method, either by maintaining similar accuracy with improved speed or by achieving higher TPR and TNR across all thresholds, along with a larger AUC. Specifically, the gray-WDLAW method outperforms the LAW method across all measures of accuracy. This improvement indicates a significantly enhanced ability to accurately identify images originating from the same source.

\begin{table}[!h]
\centering
\caption{Comparison of image source identification accuracy using DWT-based methods on the Dresden dataset.}
\label{app:result1}
\begin{tabular}{lccc}
\toprule 
\textbf{Method}  & \textbf{TPR/TNR (Youden)}  & \textbf{TPR@TNR=0.99} & \textbf{AUC} \\
\midrule 
LAW                        & 0.87/0.95  & 0.80 & 0.95 \\
gray-WDLAW                 & 0.90/0.97  & 0.87 & 0.96 \\
rgb-WDLAW                  & 0.88/0.98  & 0.84 & 0.96 \\
WDLAW-gray                  & 0.89/0.98  & 0.85 & 0.96\\
\bottomrule
\end{tabular}%
\end{table}

When considering computation time, our proposed gray-WDLAW method outperforms the original LAW method on this dataset. The average comparison time of the proposed method is noticeably faster than LAW primarily because we retain the fingerprints in the wavelet domain, which significantly reduces the number of coefficients involved in the comparison (see Section \ref{sec:oursingledwt}).

\begin{table}[!h]
\centering
\caption{Computation time (in seconds) using DWT-based methods: average fingerprint extraction time (per pair), comparison time (per pair), and total runtime.}
\label{app:result2}
\begin{tabular}{lccc}
\toprule 
\textbf{Method}  & \textbf{Extraction} & \textbf{Comparison} & \textbf{Total Time} \\
\midrule 
LAW                         & 1.127 & 0.010 & 96.700 \\
gray-WDLAW                  & 0.502 & 0.001 & 37.779 \\
rgb-WDLAW                   & 1.251 & 0.004 & 90.353 \\
WDLAW-gray                   & 1.207 & 0.004 & 86.079\\
\bottomrule
\end{tabular}%
\end{table}

A drawback is, however, observed with the rgb-WDLAW method, which exhibits the longest fingerprint extraction time. On larger datasets, this limitation could lead to substantially increased computation times. From an application perspective, although rgb-WDLAW yields high TNR and accuracy, it also increases storage requirements for the extracted fingerprints due to the three DWT computations (one for each channel). The trade-off between computation time, storage cost, and accuracy gains may not always justify its use in practical scenarios. The wavelet domain grayscaling method performs similarly to the full RGB fingerprint extraction, indicating that the additional information provided by the multichannel DWT does not bring significant benefits; we suggest that its decreased computational cost and substantially improved accuracy and subsequent the ability to identify images from the same source, makes our proposed gray-WDLAW technique a practical and realistic option for real-world company applications.

\subsubsection{DTCWT-based extraction}

The results presented in Table \ref{app:result3} and Table \ref{app:result4} correspond to the methods utilising the DTCWT. Concerning the residual image denoising method of \cite{zeng2020source}, we observe poor results in terms of accuracy, whether we include artifact removal or not when compared to the DWT-based results in Table \ref{app:result1}. We note here, however, that this approach was developed for the tampering localisation application in \cite{zeng2020source} and not fingerprint comparison specifically. Both the hybrid LAW-DTCWT and our proposed gray-WDLAW-DTCWT methods achieve high accuracy, comparable to the wavelet domain DWT-based extraction methods (see Table \ref{app:result3}).    



\begin{table}[!h]
\centering
\caption{Comparison of image source identification accuracy using DTCWT-based methods on the Dresden dataset.}
\label{app:result3}

\resizebox{\linewidth}{!}{%
\begin{tabular}{lccc}
\toprule 
\textbf{Method}  & \textbf{TPR/TNR (Youden)}  & \textbf{TPR@TNR=0.99} & \textbf{AUC} \\
\midrule 
DTCWT                                 & 0.44/0.84  & 0.08 & 0.58 \\
DTCWT with artifact removal           & 0.30/0.83  & 0.06 & 0.55 \\
LAW-DTCWT                  & 0.88/0.98  & 0.85 & 0.95 \\
gray-WDLAW-DTCWT           & 0.90/0.96  & 0.85 & 0.97 \\

\bottomrule
\end{tabular}%
}
\end{table}

As observed in \cite{zeng2020source}, the DTCWT-based methods have higher extraction times than the DWT-based methods, see Table \ref{app:result4}. This is to be expected due to the increased directional wavelet filtering (computation of the six subimages) of the DTCWT.  Our wavelet domain approach, gray-WDLAW-DTCWT, has similar performance to the hybrid approach, but is significantly faster in extraction time, sacrificing a little on the comparison time.

\begin{table}[!h]
\centering
\caption{Computation time (in seconds) using DTCWT-based methods: average fingerprint extraction time (per pair), comparison time (per pair), and total runtime.}
\label{app:result4}

\resizebox{\linewidth}{!}{%
\begin{tabular}{lccc}
\toprule 
\textbf{Method}  & \textbf{Extraction} & \textbf{Comparison} & \textbf{Total Time} \\
\midrule 
DTCWT                      & 2.775  & 0.001 & 371.310 \\
DTCWT with artifact removal                      & 3.248  & 0.001 & 434.180 \\
LAW-DTCWT                  & 2.773  & 0.001 & 372.960 \\
gray-WDLAW-DTCWT           & 1.543  & 0.004 & 233.960 \\

\bottomrule
\end{tabular}%
}
\end{table}

\subsection{VISION dataset experimental setup}
\label{sec:VISION}

To further demonstrate the advantages of the proposed WD fingerprint extraction method, we extend our experiments to a second dataset: the public VISION dataset \citep{shullani2017vision}. 

This dataset contains images captured by 35 smartphone camera devices, with each device contributing approximately 13 images. The dataset is publicly available at \textit{https://www.kaggle.com/datasets/mahmoudemam85/\\source-camera-identification-a-custom-dataset}.

Although this experimental setup does not exactly match the setting of our proposed method, it provides a valuable benchmark for evaluating the robustness and generalizability of the proposed method across a different dataset. Following the same preprocessing protocol as in the previous experiments, we centre-crop all images to the same spatial resolution (1024 $\times$ 1024) and perform exhaustive pairwise comparisons.

Based on the results from the first dataset, we observe a trade-off between computational efficiency and identification accuracy across different methods. Consequently, we focus on the grayscale first version of the proposed wavelet-domain LAW method. The grayscale in the wavelet domain implementation (WDLAW-gray) increases computational cost without yielding any measurable performance improvement. Furthermore, given the high computational complexity of the DTCWT-based extraction methods and the absence of corresponding accuracy gains, these approaches are omitted from further comparison. Therefore, in this experiment, we restrict our evaluation to the LAW, gray-WDLAW, and rgb-WDLAW methods.

\subsubsection{VISION dataset experimental results}
The results presented in Tables \ref{app:result5} and \ref{app:result6} report the identification accuracy and computational time for the three evaluated methods. As shown in the tables, even under a setting that does not strictly follow the application scenario provided by our industrial partner, the proposed method performs robustly and achieves a level of accuracy comparable to that of the existing LAW method.

Due to the larger number of images available from each device in this dataset, all methods benefit from an overall improvement in accuracy. However, the primary advantage of the proposed method becomes evident when considering computational efficiency. While achieving similar identification performance, the proposed approach requires substantially less computation time than the competing methods.

This improvement translates directly into reduced computational cost in practical industrial applications, highlighting the suitability of the proposed method for large-scale and time-sensitive deployment.

\begin{table}[!h]
\centering
\caption{Comparison of image source identification accuracy using DWT-based methods on the VISION dataset.}
\label{app:result5}
\begin{tabular}{lccc}
\toprule 
\textbf{Method}  & \textbf{TPR/TNR (Youden)}  & \textbf{TPR@TNR=0.99} & \textbf{AUC} \\
\midrule 
LAW                        & 0.96/0.99  & 0.96 & 0.98 \\
gray-WDLAW                 & 0.96/0.99  & 0.96 & 0.98 \\
rgb-WDLAW                  & 0.96/0.99  & 0.96 & 0.98 \\
\bottomrule
\end{tabular}%
\end{table}

\begin{table}[!h]
\centering
\caption{Computation time (in seconds) using DWT-based methods: average fingerprint extraction time (per pair), comparison time (per pair), and total runtime.}
\label{app:result6}
\begin{tabular}{lccc}
\toprule 
\textbf{Method}  & \textbf{Extraction} & \textbf{Comparison} & \textbf{Total Time} \\
\midrule 
LAW                         & 1.071 & 0.007 & 684.310 \\
gray-WDLAW                  & 0.507 & 0.001 & 338.398 \\
rgb-WDLAW                   & 1.214 & 0.004 & 793.302 \\
\bottomrule
\end{tabular}%
\end{table}

\subsection{Experiments on hyperparameter choice}

\subsubsection*{Wavelet resolution level parameter, $J$}
The wavelet decomposition level $J$ is an important hyperparameter in the proposed framework, as it determines both the amount of detail information retained in the extracted WD fingerprint and the corresponding computational cost. To investigate its influence, a sensitivity analysis was conducted using the proposed gray-WDLAW method with a fixed noise variance parameter $\sigma_n^2$ on the Dresden dataset. Performance was evaluated in terms of the AUC, together with the TPR and TNR under the two optimal decision criteria introduced in Section \ref{sec:Criteria}, and the results are summarised in Table~\ref{app:sensitivity1}. The results indicate that the improvement in source camera identification performance begins to plateau at resolution level $J = 3$ or $J = 4$. In particular, the results obtained with $J = 3$ and $J = 4$ are effectively identical in terms of TPR and TNR.

The choice of $J$ directly affects the size of the extracted WD fingerprint, as described in Section \ref{sec:oursingledwt}. 
Although $J=4$ was selected for the experiments in this article primarily to remain consistent with the parameter settings adopted in previous studies \cite{junior2019prnu}, the sensitivity analysis further indicates that this choice provides a favourable balance between source camera identification performance and computational efficiency. For the evaluation measures considered in this work, increasing the decomposition level beyond $J=4$ provides limited additional performance while increasing the size of the extracted fingerprint and the associated computational cost. An interesting direction of further work for our wavelet-based fingerprint extraction would be a data-driven choice of the resolution parameter, for example using cross-validation techniques used in other settings, e.g. \cite{hall2001cross, nason2002choice}.  

\begin{table}[!h]
\centering
\caption{Sensitivity analysis of $J$ on the source camera identification performance of the proposed gray-WDLAW method using the Dresden dataset ($\sigma_n^2 = 1.8^2$).}
\label{app:sensitivity1}
\resizebox{\linewidth}{!}{%
\begin{tabular}{lccc}
\toprule 
\textbf{Level $J$}  & \textbf{TPR/TNR (Youden)}  & \textbf{TPR@TNR=0.99} & \textbf{AUC}\\
\midrule 
$J = 1$                 & 0.81/0.98  & 0.75  & 0.93 \\
$J = 2$                 & 0.89/0.95   & 0.83  & 0.95 \\
$J = 3$                 & 0.90/0.97   & 0.87  & 0.96 \\
$J = 4$                 & 0.90/0.97   & 0.87  & 0.96 \\
$J = 5$                 & 0.91/0.95   & 0.87  & 0.96 \\
\bottomrule
\end{tabular}%
}
\end{table}

\subsubsection*{Noise parameter $\sigma_n^2$}
Another parameter that merits discussion here is the noise parameter $\sigma_n^2$, as it controls the shrinkage factor of the filtering applied during fingerprint extraction. In this article, for our algorithms in Section \ref{sec:Method}, the values of $\sigma_n^2$ was selected to match those used in the corresponding reference implementations for the LAW and DT-CWT methods \cite{junior2019prnu, zeng2020source}. This decision was made to ensure a fair comparison between the proposed wavelet-domain framework and the original image-domain methods, allowing the observed performance differences to be attributed to the proposed fingerprint representation rather than differences in parameter tuning. It is worth noting that the literature does not provide a universally accepted procedure for determining the optimal value of $\sigma_n^2$. Instead, previous studies typically assume a known fixed value for this parameter (and that $S_n$ is often taken to be equal to $\sigma_n$) such as $\sigma_n^2=5^2$ or $\sigma_n^2=1.8^2$, depending on the specific denoising framework.

Despite this, in this section we perform a sensitivity analysis for this noise parameter to examine its effect.  With the same Dresden dataset, as demonstrated by Table \ref{app:sensitivity2} with $J = 4$, the choice of $\sigma_n^2$ influences the source camera identification performance, with an inappropriately high value leading to reduced TPR and TNR under the evaluation criteria considered in this study.  However, for small to moderate values the algorithm is relatively robust to this choice -- a feature also observed in other PRNU studies, see e.g. \cite{lukas2006digital}.

Future work could investigate integrating adaptive (wavelet domain) noise variance estimation (such as \cite{donoho1994ideal, johnstone1997wavelet, nason2002choice, johnstone2005empirical}) into the proposed framework in and assess its impact on both identification performance and computational efficiency.

\begin{table}[!h]
\centering
\caption{Sensitivity analysis of $\sigma_n^2$ on the source camera identification performance of the proposed gray-WDLAW method using the Dresden dataset ($J = 4$).}
\label{app:sensitivity2}
\resizebox{\linewidth}{!}{%
\begin{tabular}{lccc}
\toprule 
\textbf{Method}  & \textbf{TPR/TNR (Youden)}  & \textbf{TPR@TNR=0.99} &  \textbf{AUC}\\
\midrule 
gray-WDLAW ($\sigma_n^2 = 1.8^2$) &  0.90/0.97  & 0.87   & 0.96\\
gray-WDLAW  ($\sigma_n^2 = 5^2$)  &  0.83/0.97  & 0.75   & 0.95\\
gray-WDLAW  ($\sigma_n^2 = 10^2$) &  0.77/0.91  & 0.57   & 0.90\\
\bottomrule
\end{tabular}%
}
\end{table}

\subsection{Robustness analysis}

To further evaluate the practical applicability of the proposed gray-WDLAW framework, two additional robustness experiments were conducted on the Dresden dataset. These experiments investigate the effects of common image post-processing operations and implementation choices that may be encountered in real-world forensic applications. Unless otherwise stated, the wavelet decomposition level and noise variance parameter were fixed at $J = 4$ and $\sigma_n^2=1.8^2$, respectively, consistent with the settings adopted throughout the experimental evaluation.

\subsubsection{JPEG recompression}
JPEG compression is one of the most common post-processing operations applied to digital images, for example when being uploaded to social media platforms, before they are acquired for analysis in forensic investigations \cite{rocha2011vision}. These compression algorithms tend to remove high frequency components and introduce other noise artefacts, making PRNU-based source identification more challenging \cite{goljan2016effect}. To evaluate the robustness of the proposed framework, the test images were recompressed using JPEG quality factors of 70, 85, and 95 prior to fingerprint extraction, inline with common compression settings \cite{alles2009source}. The corresponding source camera identification performance is summarised in Table~\ref{app:robust1}. The results demonstrate that whilst the detection performance decreases as expected, the proposed WD fingerprint representation remains effective under moderate JPEG compression.

\begin{table}[!h]
\centering
\caption{Robustness of the proposed gray-WDLAW method under JPEG recompression.}
\label{app:robust1}
\resizebox{\linewidth}{!}{%
\begin{tabular}{cccc}
\toprule 
 \textbf{JPEG Quality} & \textbf{TPR/TNR (Youden)}  & \textbf{TPR@TNR=0.99} &  \textbf{AUC}\\
\midrule 
70  &  0.71/0.88  & 0.41   & 0.86\\
80  &  0.80/0.88  & 0.58   & 0.92\\
95  &  0.91/0.94  & 0.83   & 0.96\\
\bottomrule
\end{tabular}%
}
\end{table}

\subsubsection{Sensitivity to cropped image size}

The proposed framework extracts fingerprints from the central region of each image. In the previous experiments presented, a crop size of $1024 \times 1024$ pixels was adopted. Some authors have noted that image cropping can degrade the fingerprint extraction, see e.g. \cite{goljan2008camera}. To investigate the robustness of the proposed framework to this design choice, additional experiments were conducted using different central crop sizes while keeping all other parameters unchanged.

The experimental results in Table \ref{app:robust2} show that larger crop sizes generally lead to improved source camera identification performance under both optimal thresholds. This behaviour is expected, as larger image regions contain more sensor pattern noise information, providing a stronger fingerprint for comparison. These results illustrate the trade-off between computational efficiency and identification performance, allowing the crop size to be selected according to the requirements of the intended forensic application.

\begin{table}[!h]
\centering
\caption{Robustness of the proposed gray-WDLAW method under various central cropping sizes.}
\label{app:robust2}
\resizebox{\linewidth}{!}{%
\begin{tabular}{cccc}
\toprule 
 \textbf{Image Size} & \textbf{TPR/TNR (Youden)}  & \textbf{TPR@TNR=0.99} &  \textbf{AUC}\\
\midrule 
$256 \times 256$  &  0.69/0.81  & 0.28   & 0.81\\
$512 \times 512$  &  0.81/0.87  & 0.54   & 0.90\\
$1024 \times 1024$  &  0.90/0.97   & 0.87  & 0.96\\
$2048 \times 2048$  &  0.94/1.00   & 0.94  & 0.98\\
\bottomrule
\end{tabular}%
}
\end{table}

\subsection{Comparison between spatial domain and WD fingerprints}
The proposed WD fingerprint is motivated by the orthonormal property of the wavelet transform, which suggests that the discriminative information contained in the transformed coefficients can be preserved without explicitly reconstructing the fingerprint into the spatial domain. However, due to the presence of non-linear operations in the proposed pipeline, including the locally adaptive Wiener-type filtering and the subsequent Fourier-domain filtering, strict mathematical equivalence of cosine similarity between the spatial domain and wavelet-domain representations is not guaranteed.

An additional empirical analysis was conducted on the Dresden dataset to investigate how closely the cosine similarity on the proposed WD fingerprint resembles that of the conventional spatial domain fingerprint. For the same set of image pairs, cosine similarity scores were independently calculated using both the conventional spatial domain fingerprint and the proposed WD fingerprint. The resulting similarity scores were then compared using 
the coefficient of determination ($R^2$).
The comparison results show a strong correlation between the two representations, with 
an $R^2$ value of 0.86. This indicates that the proposed WD fingerprint preserves a similar discriminative behaviour to the conventional spatial-domain representation for source camera identification. 

It should be noted that the objective of this analysis is not to demonstrate exact numerical equivalence between the two fingerprints, as the non-linear filtering operations are important in creating a denoiser with good properties. Instead, the experiment demonstrates that the wavelet domain representation retains the essential information required for source camera identification. The differences in classification performance underlie the lack of perfect correlation in the similarity scores.

\section{Discussion and conclusion}
\label{sec:Conclusion}

This study introduces a \textit{wavelet-domain} locally adaptive window-based denoising method that builds upon previous SPN extraction techniques. Our approach simplifies existing methodology by removing redundant steps that could degrade fingerprint detection accuracy and introduce unnecessary computational cost when performing SPN comparisons. In practical applications, where security agencies may need to identify cameras from vast collections of images, the computational cost of wavelet inversion can contribute significantly to computation. By eliminating the need for the inversion, our proposed gray-WDLAW technique offers a more efficient solution, reducing processing time while maintaining the integrity of the fingerprinting process. Experimental evaluations demonstrate that 

the proposed method achieves detection accuracy comparable to, and in some configurations marginally superior to, the conventional LAW pipeline (AUC 0.97 vs. 0.95 on the Dresden dataset). The primary advantage, however, lies in computational efficiency rather than a substantial accuracy gain.

Further comparisons show that the proposed method is versatile, working effectively not only with the DWT but also with other more recent wavelet decomposition techniques such as the DTCWT. Our work demonstrates that, depending on the application or researchers' needs, performance and/or efficiency improvements can be gained by adopting our general approach.

Future work could explore alternative similarity metrics beyond cosine similarity, which was chosen for its simplicity and computational efficiency. Binary Hamming distance may serve as an efficient alternative for similarity testing, particularly for data storage or compression purposes \cite{bayram2012efficient}.

In addition, we suggest that wavelet domain approaches may be able to be combined with other strategies for efficient storage and comparison of PRNU fingerprints, such as those introduced in \cite[Section VII]{al2016spn}. Beyond Mihcak’s filters, other filtering techniques \cite{argenti2006mmse, amerini2009analysis, cortiana2011performance} could also be integrated into the WDLAW framework: since our focus is on structural modifications rather than specific filtering methods, users are free to explore and incorporate any filter that fits within this structure.

\section*{Data Availability Statement}

The datasets used for this study are available from the links provided in the text of Section \ref{sec:DWT} and \ref{sec:VISION}.

\bibliographystyle{elsarticle-harv}
\bibliography{references}

\end{document}